\documentclass[letterpaper,10pt,journal,twoside]{IEEEtran}  

\usepackage{cite}
\usepackage{amsmath,amssymb,graphicx}
\usepackage{xcolor}
\usepackage{soul}
\usepackage{graphicx}
\usepackage[percent]{overpic}
\usepackage{subcaption}
\usepackage{tikz}
\usepackage{wrapfig}
\usepackage{booktabs}
\usepackage[hidelinks]{hyperref}
\title{\LARGE \bf
HALOMI: Learning Humanoid Loco-Manipulation with Active Perception from Human Demonstrations
}

\author{%
Zehui Zhao$^{1,*}$, Yuxuan Zhao$^{1,*}$, Gaojing Zhang$^{1,2}$, 
Chenxi Liu$^{3}$, Maolin Zheng$^{3}$, Wenzhao Lian$^{1,\dagger}$\\
{\normalsize $^{1}$Shanghai Jiao Tong University}
{\normalsize $^{2}$University of Sussex}\\
{\normalsize $^{3}$East China University of Science and Technology}\\
{\normalsize $^{*}$Equal Contribution, $^{\dagger}$Corresponding Author}
}

\begin{document}


    

\IEEEaftertitletext{%
\vspace{-3em}
\begin{center}
    \includegraphics[width=0.96\textwidth]{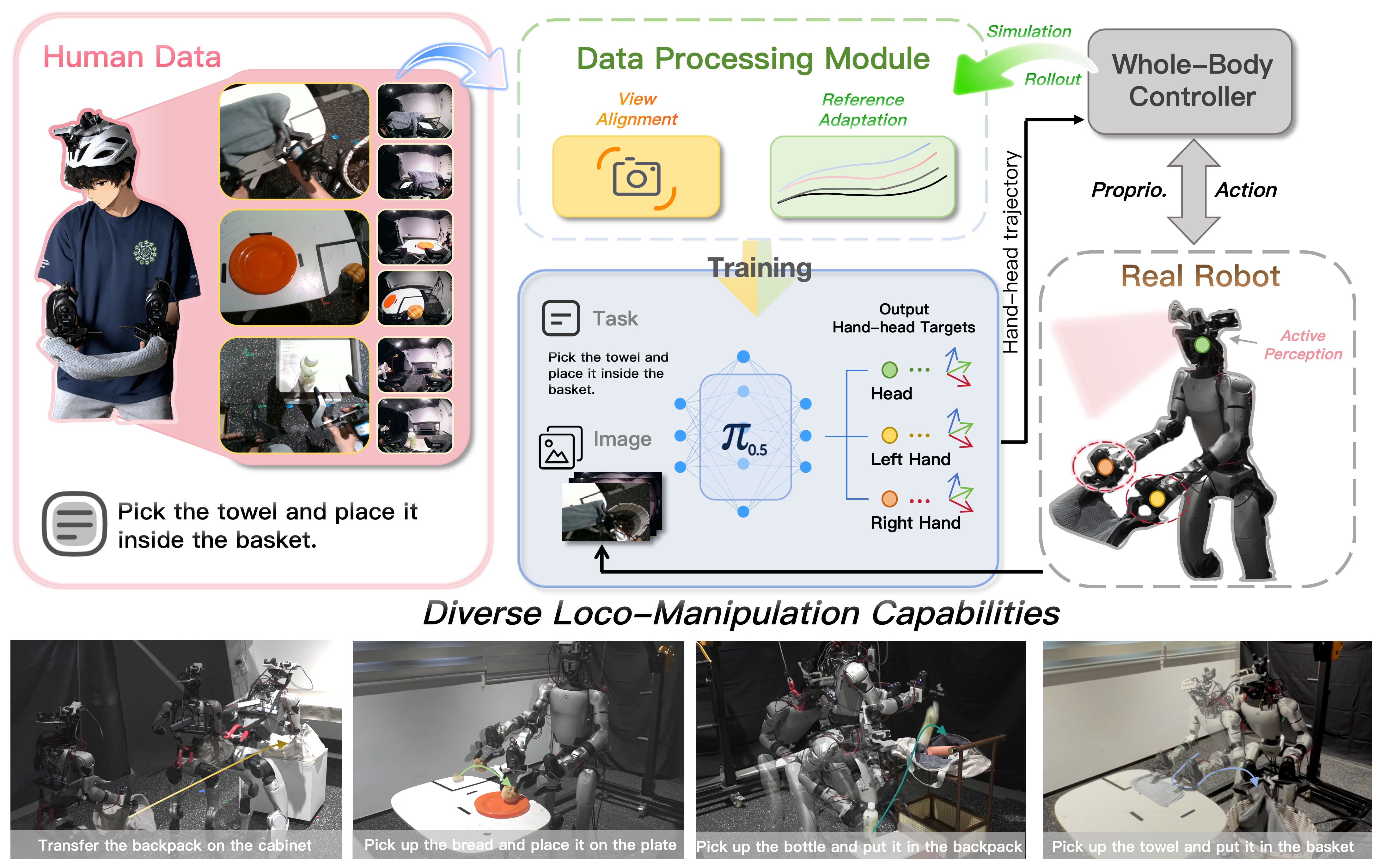}

    \vspace{0.35em}
    \refstepcounter{figure}
    \label{fig:teaser}
    \parbox{0.96\textwidth}{%
    \normalfont\normalsize
    \textbf{Fig.~\thefigure.}
    \textbf{Humanoid Active-Perception Loco-Manipulation Interface (HALOMI).} Human demonstrations are collected with UMI-style handheld grippers and wearable egocentric sensing.
    The data processed by the human data processing module are used to train a high-level VLA that predicts head-hand targets. The head-hand targets are then executed by a manifold-constrained whole-body controller on the humanoid robot.
    }
\end{center}
\vspace{-0.4em}
}

\maketitle
\thispagestyle{empty}
\pagestyle{empty}

\begin{abstract}
Human demonstrations, which can be collected at scale and naturally capture active hand-eye coordination, are a promising data source for learning humanoid loco-manipulation. However, directly transferring human demonstrations to humanoids requires a precise world-frame tracking controller, which is often brittle under Out-of-Distribution(OOD) targets, while human-to-humanoid gaps persist in both egocentric observation and action execution. 
To address these challenges, we present HALOMI, a scalable framework for learning humanoid loco-manipulation with active perception from human demonstrations.
HALOMI extends Universal Manipulation Interface (UMI) with egocentric sensing to collect ego-view and wrist-view observations along with head-hand trajectories at scale.
We further propose a manifold-constrained controller that plans in a learned latent behavior manifold to enable precise and robust head-hand tracking in the world frame. To bridge the human–to-humanoid gap, we perform ego-view alignment and introduce a controller-aware reference trajectory adaptation to reduce mismatch in both observation and action execution. We validate HALOMI on a Unitree G1 humanoid robot with an actuated neck across five real-world tasks involving navigation, grasping, bimanual manipulation, whole-body coordination, and dynamic behaviors. Across the three quantitatively evaluated tasks, HALOMI achieves an average success rate of 85\%, while additional qualitative demonstrations show its ability to support dynamic tossing and deep-squat grasping. Our project website can be found at \href{https://halomi-humanoid.github.io}{https://halomi-humanoid.github.io}

\end{abstract}

\section{INTRODUCTION}





Humanoid robots, with their human-like morphology and capacity for flexible whole-body coordination, are promising robotic platforms for operating in unstructured human-centric environments. However, collecting large-scale teleoperated demonstrations for humanoids remains difficult, as whole-body teleoperation is costly, time-consuming, and requires access to physical humanoid platforms as well as skilled operators.



Meanwhile, human demonstrations provide an appealing alternative for scaling humanoid loco-manipulation vision-language-action (VLA) models, as they can be collected without costly real-robot teleoperation. Beyond scalability, human demonstrations are ego-view-guided action data. Humans coordinate head and hand motions to maintain task-relevant visual context, enabling long-horizon tasks with multi-stage gaze changes such as search, grasping, and placement\cite{vision_in_action}. Since the demonstrated actions are generated under actively changing ego-view observations, active perception is also required to transfer such data to humanoids.


Robot-free demonstration frameworks have combined UMI-style handheld interfaces~\cite{umi} with egocentric sensing to collect human demonstrations that capture hand-object interactions and active viewpoint behaviors~\cite{activeumi,hommi,egomi}. These systems demonstrate the value of human demonstrations for learning manipulation tasks that require hand-eye coordination, yet they primarily focus on fixed-base or wheeled robotic platforms.


Recent works have extended robot-free demonstrations to humanoid loco-manipulation by augmenting UMI interfaces with additional lower-body references, such as pelvis or foot targets~\cite{humi,bifrostumi}.
These additional references help specify humanoid whole-body motion, but require access to lower-body information during data collection.
Moreover, these systems do not study active ego-view transfer from humans to humanoids, limiting their applicability to long-horizon tasks that require continual gaze changes.
%


Motivated by these limitations, we adopt a task-centric head-hand demonstration interface that captures manipulation-relevant hand motions and active head movements, while a whole-body controller completes the whole-body motion required for stable execution.
However, this interface introduces additional challenges, including (1) the ego-view observation gap between humans and humanoids due to morphology and camera-pose differences, (2) directly executing world-frame head-hand trajectories with unspecified lower-body commands is brittle and unstable for the whole-body controller, (3) human demonstrations provide desired head-hand target states, while directly feeding these targets to the humanoid whole-body controller can yield non-negligible tracking errors, thereby degrading human-to-humanoid transfer.

To address these challenges, we present the \textbf{H}umanoid \textbf{A}ctive-Perception \textbf{Lo}co-\textbf{M}anipulation \textbf{I}nterface (\textbf{HALOMI}), as illustrated in Fig.~\ref{fig:teaser}, a general and scalable framework for \textbf{learning humanoid loco-manipulation with active perception from human demonstrations}. HALOMI includes a \textbf{UMI-augmented egocentric data collection} protocol, a \textbf{manifold-constrained whole-body controller} for robust and precise tracking of head-hand targets, and an \textbf{automated human data processing procedure}  to reduce the human-to-humanoid gap. The main contributions of our system are summarized as follows:

\begin{itemize}
\item\textbf{HALOMI Data Collection System:} We develop a robot-free egocentric data collection system that combines bimanual UMI-style handheld grippers with a wearable head-mounted egocentric sensing, enabling synchronized collection of multi-view visual observations and corresponding hand-eye motion trajectories.
\item\textbf{Manifold-Constrained Whole-Body Controller:} We develop a manifold-constrained reinforcement learning (RL) controller that tracks a unified set of head-hand world-frame targets to bridge the high-level VLA action interface and humanoid whole-body execution. Instead of tracking in the joint action space, the controller plans over a learned latent behavior manifold, enabling precise and robust world-frame tracking.
\item\textbf{View Alignment \& Reference Trajectory Adaptation:} We introduce an automated offline data processing pipeline to reduce the human-to-humanoid embodiment gap in both observation and action spaces. View alignment mitigates visual discrepancies caused by viewpoint mismatch, while controller-aware reference trajectory adaptation reduces execution errors of human head-hand trajectories, mitigating error accumulation during closed-loop rollouts.

\item\textbf{Hierarchical Learning Framework for Human-to-Humanoid Transfer:} We train a high-level VLA policy from scalable human demonstrations processed by the automated data processing pipeline. The VLA policy takes ego-view and dual-hand visual observations as input and predicts head-hand targets, which are then executed by the manifold-constrained whole-body controller for humanoid execution. We conduct real-world experiments on a Unitree G1 humanoid across diverse loco-manipulation tasks, demonstrating effective human-to-humanoid skill transfer.

\end{itemize}

\section{RELATED WORK}

\subsection{Robot-Free Human Demonstrations} 


Robot-free demonstrations have emerged as a scalable pathway for robot policy learning without relying on real-robot teleoperation. UMI-style interfaces have shown effective transfer to fixed-base manipulators~\cite{umi,umift,dexumi} and have been extended to mobile embodiments~\cite{umi_on_leg,humi,hommi,bifrostumi}. For robot-free human demonstrations with active perception, egocentric sensing captures not only visual observations, but also human head motions that actively reveal task-stage-specific context during long-horizon manipulation~\cite{activeumi,egomi}.
HALOMI builds on these directions by augmenting UMI-style hand-trajectory collection with wearable egocentric head sensing, capturing both hand-object interaction and active viewpoint behavior for humanoid loco-manipulation.

\subsection{Humanoid Whole-Body Controller} 
Recently, RL-based controllers for humanoid robots have achieved impressive sim-to-real performance, enabling humanoids to perform diverse loco-manipulation tasks. HOMIE\cite{homie} and AMO\cite{amo} adopt a decoupled humanoid controller, where upper-body motion is specified by joint-space targets and the RL controller handles lower-body locomotion and balance. 
Another line of work builds on the motion-tracking paradigm\cite{twist,omnih2o}, enabling humanoids to reproduce human-like whole-body motions. Recent systems\cite{clone,clot} operate in the global frame to reduce global drift during teleoperation. In addition, motion-prior and latent-space controllers\cite{uhc,bfmzero,Li_cvpr} have been explored to constrain whole-body control within a learned behavior manifold, improving motion feasibility compared with unconstrained joint-space tracking.

Executing recorded human head-and-hand trajectories requires precise world-frame sparse keypoint tracking, which is sensitive to localization errors and infeasible commands. Since pelvis, foot, and lower-body references are not provided, the controller must infer the remaining whole-body motion from only head-hand targets. Together, these factors make the sparse world-frame tracking challenging. HALOMI therefore adopts the manifold-constrained controller, which tracks the head-hand targets by planning over a learned latent behavior manifold, enabling precise and robust world-frame tracking.


\begin{figure}[tp]
    \centering

    \begin{overpic}[
        width=0.93\columnwidth,
        height=0.65\columnwidth
    ]{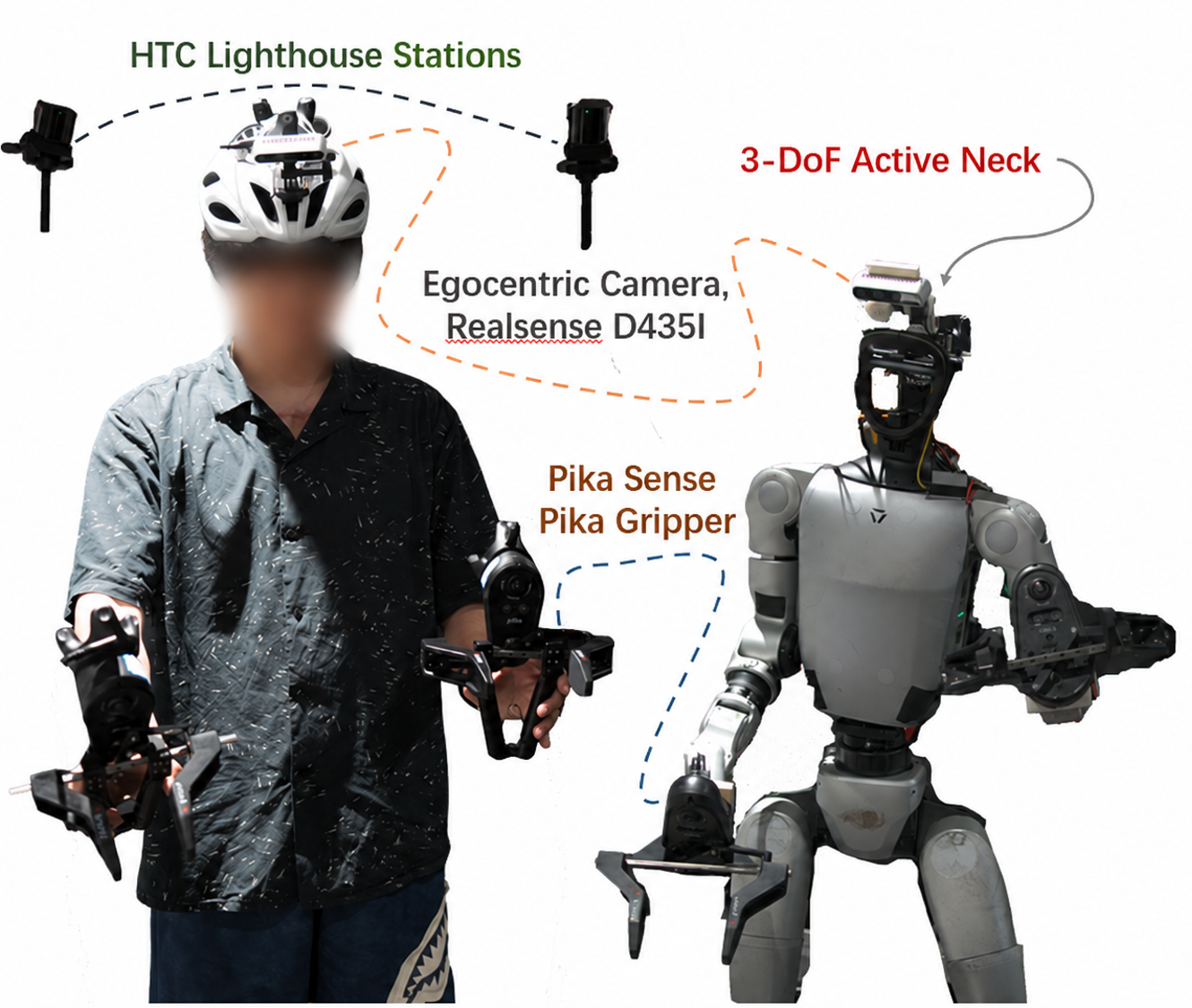}
        \put(24,-2.6){\small\bfseries (a)}
        \put(72,-2.6){\small\bfseries (b)}
    \end{overpic}

    \vspace{0.3em}

    \caption{
    \textbf{Hardware overview of the proposed system:} (a) human data collection setup and (b) robot hardware platform.
    }
    \label{fig:data_collect}
     \vspace{-5mm}
\end{figure}

\subsection{Whole-Body Policy Learning from Human Demonstrations} 
EgoHumanoid\cite{egohumanoid} attempts to transfer the human egocentric data to the humanoid robot via view and action alignment and human-robot co-training. HoMMI~\cite{hommi} integrates a 3D egocentric representation with a look-at-point head interface to mitigate the embodiment gap for whole-body mobile manipulation. 


HuMI~\cite{humi} and BifrostUMI~\cite{bifrostumi} extend UMI-style robot-free demonstrations to humanoid loco-manipulation by augmenting hand trajectories with additional lower-body references, such as pelvis and foot targets. 
These additional references help specify humanoid whole-body motion, but they also require demonstrators to provide lower-body information and rely on morphology-aware adaptation or retargeting for execution. 
Since pelvis and foot targets lack direct visual anchors, VLA-predicted lower-body references can be prone to long-horizon drift.
Moreover, both systems do not explicitly include active perception, limiting their applicability to long-horizon tasks that require continual gaze changes. 
In contrast, HALOMI uses head-hand targets as a task-centric and embodiment-agnostic interface. 
A unified manifold-constrained whole-body controller executes these sparse targets and completes the remaining whole-body motion, while an automated data-processing pipeline reduces human-to-humanoid gaps in egocentric observation and action execution. 
These designs aim to reduce the burden of data collection and 
enable robot-free human demonstrations with active gaze behaviors to serve as a scalable source for learning humanoid loco-manipulation skills.

\section{PROPOSED FRAMEWORK}

HALOMI consists of four main components, including a scalable and intuitive human data collection system paired with a humanoid platform (Sec. \ref{sec:dat_collect}), a unified whole-body RL controller for precise and robust head-hand target tracking (Sec. \ref{sec:controller}), an automated offline alignment pipeline for processing raw human demonstrations (Sec. \ref{sec:alignment}), and the high-level VLA training and deployment pipeline (Sec. \ref{sec:vla}).

\subsection{Data Collection \& Robot System} \label{sec:dat_collect}

\begin{figure}[t]
    \centering
    \includegraphics[
        width=0.9\columnwidth,  
        keepaspectratio          
    ]{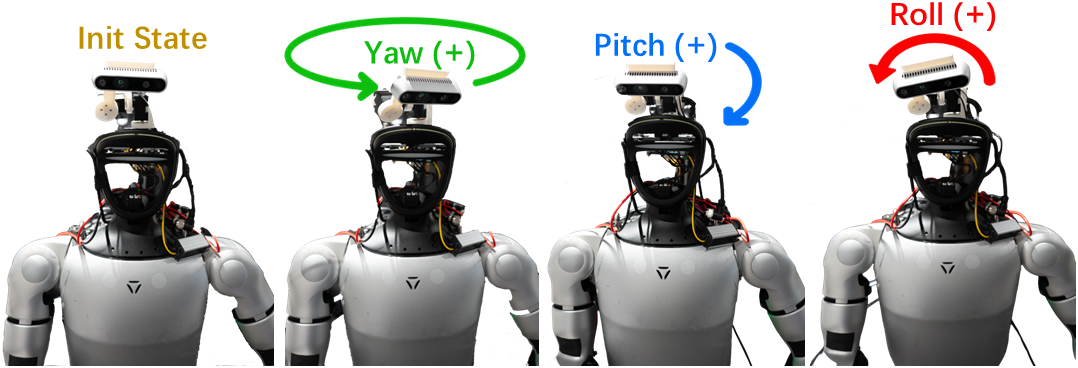}
    \caption{
    \textbf{Illustration of the 3-DoF active neck motion:} initial state, positive yaw, positive pitch, and positive roll.
    }
    \label{fig:active_neck_motion}
     \vspace{-6mm}
\end{figure}

\textbf{Scalable and Precise Robot-Free Demonstration.} As shown in Fig.~\ref{fig:data_collect}(a), our data collection system integrates bimanual UMI-style~\cite{umi} handheld grippers with wearable egocentric sensing. The demonstrator manipulates objects using two Agilex Robotics Pika Sense handheld grippers while wearing a helmet equipped with an Intel RealSense D435i camera and an additional HTC VIVE Tracker 3.0. With external Lighthouse base stations, the system tracks the relative 6-DOF trajectories of both grippers and the head with millimeter-level accuracy, while each gripper records a continuous gripper-width signal. Together, these trajectories and gripper states define the unified human-robot action space. The observation space consists of two local gripper-centric RGB streams from the Pika Sense fisheye cameras and one global egocentric RGB stream from the helmet-mounted RealSense camera. All observation and action streams are synchronized at 30 Hz.

\begin{figure*}[t]
    \centering
    \includegraphics[width=\linewidth]{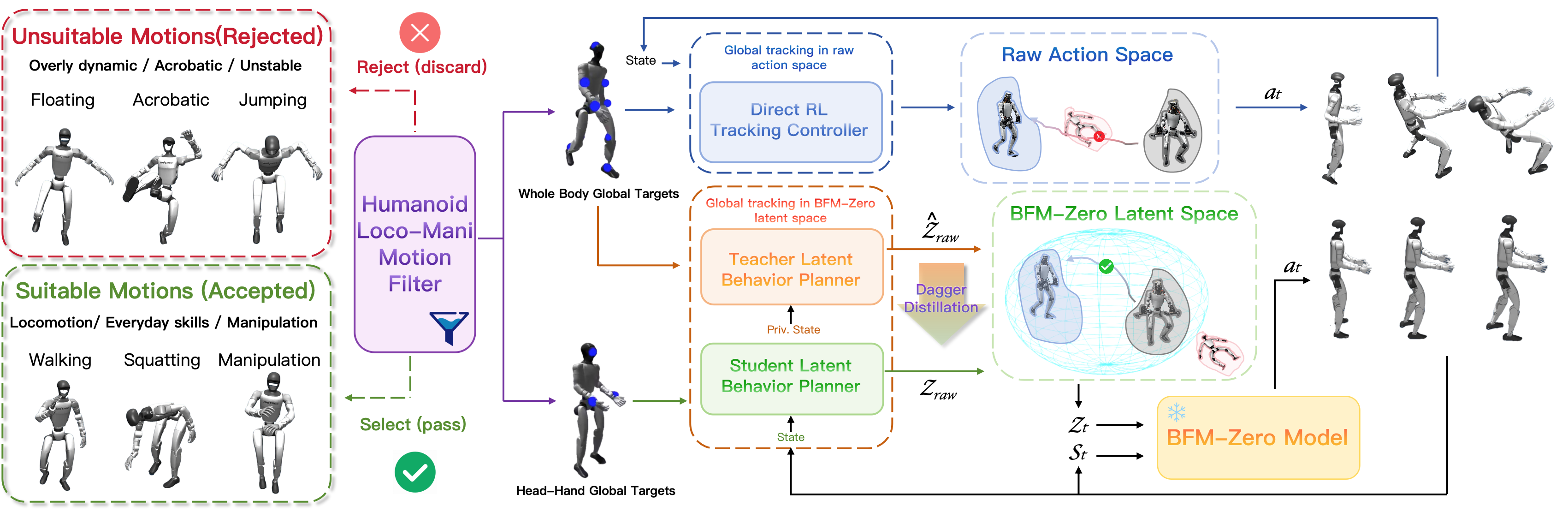}
    \caption{
    \textbf{Manifold-Constrained Whole-Body Controller} tracks sparse world-frame head–hand targets by predicting latent actions in the BFM-Zero action space. These latent actions are decoded by the BFM-Zero model into feasible whole-body actions, constraining humanoid execution to physically plausible loco-manipulation behaviors. In contrast, directly training RL for sparse world-frame tracking in the raw action space is under-constrained and may lead to aggressive and unstable motions.
    }
    \label{fig:loco}
     \vspace{-5mm}
\end{figure*}

\textbf{Robot Hardware with Active Neck.} As depicted in Fig.~\ref{fig:data_collect}(b), we use the Unitree G1 humanoid robot as the whole-body loco-manipulation platform and augment it with robot-mounted Pika grippers and a self-designed 3-DoF active neck. The robot-mounted Pika grippers share similar gripper geometry, hardware layout, and camera placement with the handheld Pika Sense devices used for human data collection.
This mechanically and visually aligned design reduces the gripper-centric observation gap between human demonstrations and humanoid policy execution.

To enable active perception on the humanoid, we further design a servo-driven 3-DoF active neck and mount it on the head of the Unitree G1. 
The neck is designed such that the three rotational axes are approximately aligned with the optical center of the RGB camera. In addition, the intermediate links between adjacent servo joints are made as short as possible, making their translational offsets negligible in practice. Hence, the active neck can be modeled as a compact 3-DoF rotational mechanism that directly controls the yaw, pitch, and roll of the robot's head-mounted camera, as demonstrated in Fig. \ref{fig:active_neck_motion}.

\subsection{Manifold-Constrained Whole-Body Control System} \label{sec:controller}

To enable scalable and transferable human data collection, we adopt a sparse manipulation-aware interface that records only the head and two hands, without torso, leg, or foot references. As a result, the remaining whole-body motion is under-specified during execution, requiring the controller to infer feasible whole-body motion from sparse head-hand targets. We therefore train a unified whole-body controller that tracks arbitrary feasible three-point world-frame targets while maintaining natural and stable whole-body motion.



Our whole-body control system, which aims to track the head and hand target poses, consists of two modules: an RL-based humanoid body controller and a 3-DoF active-neck controller with analytical inverse kinematics. Given the head-hand targets, the body controller tracks the 6-DoF poses of both hands and the 3D position of the head, while the neck controller converts the head orientation target into 3-DoF neck joint commands.

This neck–body decoupled design is motivated by the kinematic limitations of the original Unitree G1, which lacks an independently actuated neck. If the whole-body policy directly tracks head orientation, small head rotation changes may require large torso, pelvis, or leg motions, which can compromise whole-body stability and hand-tracking accuracy.

As illustrated in Fig.~\ref{fig:loco}, directly learning a precise world-frame tracking policy in the raw joint action space is brittle. World-frame tracking is inherently multi-modal: given the same target, the policy can realize it with many different solutions, including aggressive motions. These motions are often unstable, unnatural, and unsafe, but they may receive higher tracking rewards because they reduce tracking errors more quickly. This issue becomes more serious in the sparse world-frame keypoint tracking setting. To address this, we formulate world-frame tracking as planning on a learned behavior manifold rather than directly producing raw joint commands.

Specifically, we use BFM-Zero~\cite{bfmzero} as a learned behavior prior for realizing world-frame tracking. BFM-Zero learns a spherical latent behavior space, where each latent action can be decoded by the BFM-Zero model into a natural and stable whole-body motion. We therefore train an RL tracking policy in the BFM-Zero latent space to realize sparse world-frame tracking.

To train a unified head-hand world-frame tracking controller, we curate a large-scale loco-manipulation motion dataset containing over 6,000 motion sequences and adopt a teacher-student training strategy. We first train a teacher policy via RL with access to future reference motions, whole-body global and local tracking targets, and privileged information. We then distill a deployable student policy via DAgger, which only observes head-hand targets and proprioception available on the real robot. 

\textbf{Teacher Policy Training}. 
The teacher policy \(\pi_{\mathrm{tea}}\) is trained via RL with task tracking observations, privileged simulation states, and future reference motion:
\(
o_t^{\mathrm{tea}} =
\left[
o_t^{\mathrm{task}},
o_t^{\mathrm{priv}},
o_t^{\mathrm{future}}
\right].
\) Here \(o_t^{\mathrm{priv}}\) denotes privileged states available only during teacher training.
The teacher task observation contains whole-body reference targets and tracking errors:
\(
o_t^{\mathrm{task}} =
\big[
\Delta p_t^{\mathrm{root}},
v_t^{\mathrm{root}},
R_t^{\mathrm{root}},
\omega_t^{\mathrm{root}},
\hat q_t,
\Delta p_t^{\mathrm{body}},
\Delta R_t^{\mathrm{body}}
\big].
\)
where \(\Delta p_t^{\mathrm{root}}\) denotes the world-frame root position tracking error, \(v_t^{\mathrm{root}}\), \(R_t^{\mathrm{root}}\), and \(\omega_t^{\mathrm{root}}\) denote the root linear velocity, root orientation, and root angular velocity, respectively. \(\hat q_t\) denotes reference joint positions and \(\Delta p_t^{\mathrm{body}}\), \(\Delta R_t^{\mathrm{body}}\) denote the world-frame position and orientation tracking errors of the whole-body keypoints. 
The future term is
\(
o_t^{\mathrm{future}}=
\left\{
o_{t+\delta}^{\mathrm{task}}
\right\}_{\delta\in\{5,10,\ldots,50\}}.
\)
The teacher first outputs a 128-D latent command, 
\(z_t^{\mathrm{tea}}=\pi_{\mathrm{tea}}(o_t^{\mathrm{tea}})\),
which is projected onto the BFM-Zero spherical latent space as \(\tilde{z}_t^{\mathrm{tea}}\) and 
decoded by the frozen BFM-Zero policy into joint commands
\(a_t^{\mathrm{tea}}=\pi_{\mathrm{BFM}}(\tilde{z}_t^{\mathrm{tea}},s_t^{\mathrm{bfm}})\).
The training rewards include tracking terms for hand poses and head position, together with regularization terms to encourage smooth behaviors.

\textbf{Deployable Student Policy Distillation}. 
The student policy \(\pi_{\mathrm{stu}}\) observes the deployable proprioceptive history and head-hand task observations:
\(
o_t^{\mathrm{stu}} =
\left[
s_{t-10:t}^{\mathrm{stu}},
o_t^{\mathrm{task}}
\right].
\)
The student task observation contains only world-frame head-hand tracking errors:
\(
o_t^{\mathrm{task}} =
\left[
\Delta p_t^{\mathrm{head}},
\Delta p_t^{\mathrm{lh}},
\Delta p_t^{\mathrm{rh}},
\Delta R_t^{\mathrm{lh}},
\Delta R_t^{\mathrm{rh}}
\right].
\)
The student predicts a 128-D latent command
\(
z_t^{\mathrm{stu}}=\pi_{\mathrm{stu}}(o_t^{\mathrm{stu}}),
\)
which is projected onto the spherical latent space and then decoded by the frozen BFM-Zero policy:
\(
a_t^{\mathrm{stu}}=\pi_{\mathrm{BFM}}(\tilde{z}_t^{\mathrm{stu}},s_t^{\mathrm{bfm}}).
\)
We optimize the student policy through DAgger:
\begin{equation}
\mathcal{L}_{\mathrm{stu}}
=
\left\|
\pi_{\mathrm{stu}}(o_t^{\mathrm{stu}})
-
\pi_{\mathrm{tea}}(o_t^{\mathrm{tea}})
\right\|_2^2 
\end{equation}



\subsection{Human Data Processing to Bridge the Embodiment Gap} \label{sec:alignment}
Directly training an end-to-end high-level VLA policy on raw human demonstrations can lead to transfer failures during deployment, as human-to-humanoid gaps exist in both ego-view observations and action execution.
 We therefore present a two-step data processing module to ease the embodiment gap, including (1) ego-view alignment, which reduces visual mismatch by transforming human observations toward the humanoid viewpoint, and (2) Controller-aware reference trajectory adaptation adjusts the collected head-hand trajectories to enable the whole-body controller to better reproduce the demonstrated motions during humanoid execution.


\subsubsection{Ego-view Alignment}
We adopt the ego-view alignment procedure from EgoHumanoid \cite{egohumanoid}. Specifically, we first reconstruct the human egocentric observation into a 3D scene representation by a learning-based single-view depth estimation model, then reproject it to the humanoid viewpoint, and use image inpainting to fill missing regions caused by reprojection and disocclusion.

\begin{figure}[t]
    \centering
    \includegraphics[width=\linewidth]{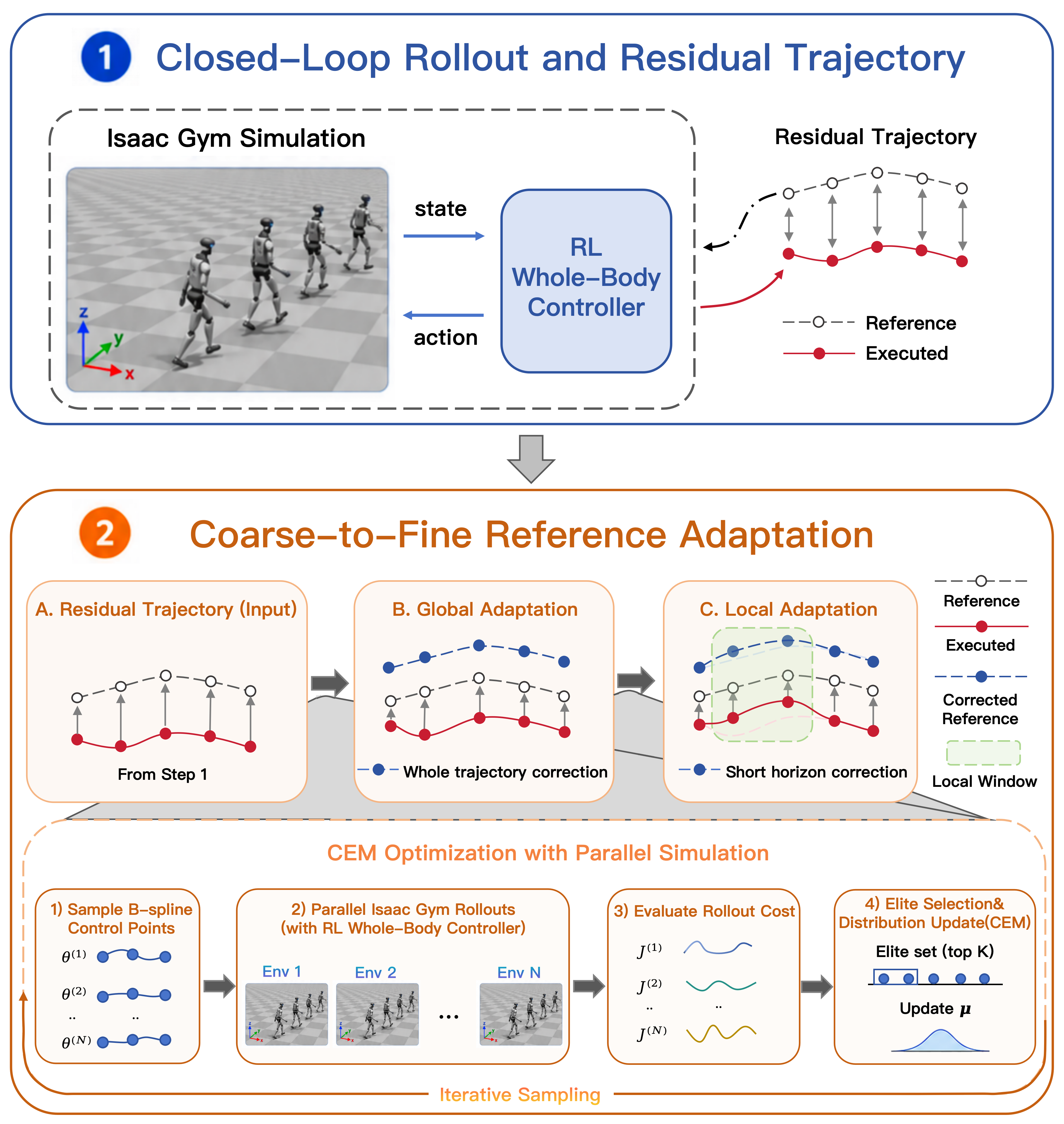}
    \caption{
    \textbf{Controller-Aware Reference Trajectory Adaptation Overview}. We first obtain the tracking errors by rolling out the reference with the whole-body controller, then perform coarse-to-fine global and local adaptation with parallel simulation.
    }
    \label{fig:ref_refinement}
    \vspace{-5mm}
\end{figure}

\subsubsection{Controller-Aware Reference Trajectory Adaptation}
As illustrated in Fig.~\ref{fig:ref_refinement}, the goal of controller-aware reference trajectory adaptation is to adjust the raw human-derived reference trajectory so that, after closed-loop execution by the humanoid whole-body controller, the executed robot trajectory better reproduces the collected human reference trajectory.

The collected human demonstration is converted into a head-hand reference trajectory
anchored at a fixed start pose:
\(
\mathbf{x}^{\mathrm{ref}}_t =
\left[
\mathbf{p}^L_t,\;
\mathbf{p}^R_t,\;
\mathbf{p}^H_t,\;
\mathbf{R}^L_t,\;
\mathbf{R}^R_t
\right].
\)
Here, \(\mathbf{p}^L_t,\mathbf{p}^R_t,\mathbf{p}^H_t\) denote the target positions of the
left hand, right hand, and head, and \(\mathbf{R}^L_t,\mathbf{R}^R_t\) denote the target
orientations of the two hands.


In practice, we adapt the translational components 
\(\mathbf{p}^{\mathrm{ref}}_t=[\mathbf{p}^L_t,\mathbf{p}^R_t,\mathbf{p}^H_t]\) of the reference trajectory through a residual trajectory~$\mathbf{r}_t$, which is parameterized by the B-splines. The adapted reference is 
 \(
\tilde{\mathbf{p}}_t = \mathbf{p}^{\mathrm{ref}}_t + \mathbf{r}_t
\). 
To obtain \(\mathbf{r}_t\), we first roll out the raw reference trajectory
in simulation from the same fixed start pose, and compute an initial residual trajectory which serves as the starting point of the subsequent multi-scale adaptation:
\(
\mathbf{e}^t = \mathbf{p}^{\mathrm{ref}}_t - \mathbf{p}^{\mathrm{exec}}_t,
\mathbf{r}^{0}_t = \Pi(G \mathbf{e}^t)
\), where \(G\) is a diagonal compensation gain and \(\Pi\) projects the correction to a feasible range. 

To correct the reference trajectory at different temporal scales, we decompose \(\mathbf{r}_t\) into a full-horizon residual component and a local residual component: \(
\mathbf{r}_{1:T}
=
\mathbf{r}^{\mathrm{global}}_{1:T}
+
\mathbf{r}^{\mathrm{local}}_{1:T}
\).
We adopt a coarse-to-fine strategy for the full-horizon global adaptation, and then further apply chunk-wise local adaptation over short temporal windows.

We first fit an initial set of global B-spline control points
\(\mathbf{C}_{\mathrm{global}}^{0}\) from \(\mathbf{r}_{1:T}^{0}\).
After global adaptation, local adaptation is performed over short temporal
windows \(\mathcal{W}_m\), each initialized from its corresponding residual
within that window: 

\begin{equation}
\begin{aligned}
\mathbf{r}_{\mathcal{W}_m}^{\mathrm{local},0}
&=
\left(
\mathbf{r}_{1:T}^{0}
-
\mathbf{r}_{1:T}^{\mathrm{global}}
\right)_{\mathcal{W}_m}, \\
\mathcal{W}_m
&=
[m,\, m+T_m-1]
\end{aligned}
\end{equation}

\begin{figure*}[t]
    \centering
    \includegraphics[
        width=0.98\textwidth,
        keepaspectratio
    ]{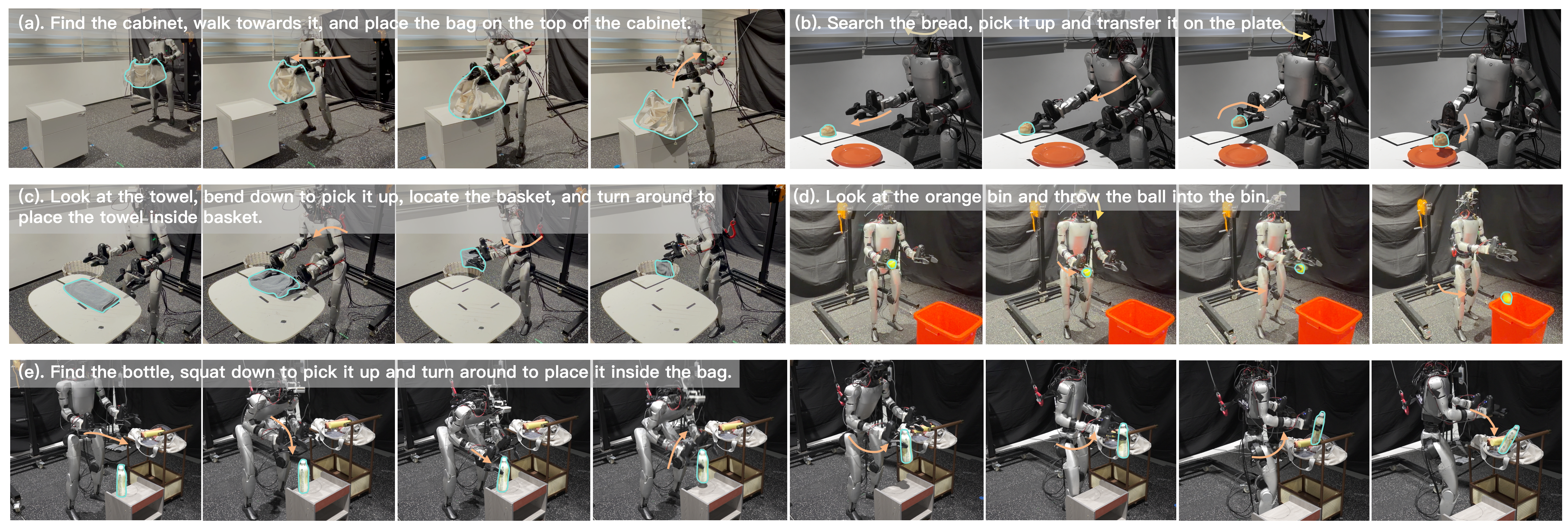}
    \vspace{-0.6em}
    \caption{
    \textbf{Real-world task setup.}
    We evaluate HALOMI on five diverse humanoid loco-manipulation tasks involving navigation, hand-eye coordination, active perception, and dynamic interaction. 
    The task instruction and sub-stage are overlaid on each policy rollout sequence, and task-relevant objects or motion directions are highlighted with visual markers for better visualization.
    }
    \label{fig:real_world_tasks}
    \vspace{-5mm}
\end{figure*}

These global and local control points serve as the initialization of the
subsequent optimization. Since the controller-simulation loop is non-differentiable, we further optimize these control points using CEM-based\cite{cem} sampling.

For the global and local adaptation stages, we optimize the B-spline control
points with CEM-based sampling. At each iteration, we sample \(K\) candidate control-point sets \(\{\mathbf{C}^{(i)}\}_{i=1}^K\) and evaluate them in
parallel using \(K\) simulation environments. For each sampled
candidate, we reconstruct the corresponding corrected trajectory, convert it into delta commands, and execute the resulting candidate via the controller in simulation, and score it by
\begin{equation}
J^{(i)} = \sum_{t=1}^{T}
\left(
\lambda_{\mathrm{track}}
\|\mathbf{p}_{t}^{\mathrm{exec},(i)}-\mathbf{p}_{t}^{\mathrm{ref}}\|
+\lambda_{\mathrm{reg}}\mathcal{R}_{t}^{(i)}
\right)
\end{equation}
Here, \(\mathcal{R}_t^{(i)}\) regularizes the smoothness of the executed trajectory.
The sample distribution is then updated from the elite set \(\mathcal{E}\)
with the lowest rollout costs:
\begin{equation}
\mathbf{C}^{(i)} \sim \mathcal{N}(\mu,\Sigma), \qquad
\mu \leftarrow \frac{1}{|\mathcal{E}|}
\sum_{\mathbf{C}^{(i)}\in\mathcal{E}} \mathbf{C}^{(i)} 
\end{equation}

Finally, we replay the adapted reference trajectories over the full horizon and select the one with the best execution score, accepting it only when it improves over the original reference trajectory.

\subsection{Loco-Manipulation Policy Learning and Deployment} \label{sec:vla}

We use the processed human data to fine-tune $\pi_{0.5}$~\cite{pi0.5} as the high-level policy.
This high-level VLA takes all synchronized RGB images and the corresponding task prompts, and predicts the action chunk.
Following FastUMI~\cite{fastumi}, we represent each action chunk as relative trajectories for three tracking targets: the left hand, right hand, and head.
Thus, we slice the human dataset and form the action chunk, for which the k-th relative pose inside the action chunk of the timestamp t for each target can be expressed as
\begin{equation}
    p_t^k = p_{t+k} - p_t, \quad R_t^k = R_t^{-1}R_{t+k}
    \label{eq:action_chunk}
\end{equation}

where the $p$ and $R$ denote the position and rotation parts of the pose. Both continuous gripper values from the encoders of the left and right Pika Sense are appended to the back of their relative pose. 

Fig.~\ref{fig:teaser} illustrates the deployment pipeline of HALOMI. At inference time, we launch an asynchronous image acquisition thread that synchronizes all camera streams at 30 Hz and continuously updates an image buffer with the latest time-stamped observations. The high-level VLA queries the buffer for the most recent synchronized egocentric and gripper-centric images together with the language instruction. It then predicts an action chunk consisting of relative pose commands for the left hand, right hand, and head, together with the gripper control signals. Each relative pose command is anchored at the corresponding absolute pose and converted into a sequence of world-frame Cartesian targets. These head-hand targets are streamed to the whole-body controller, which runs at 50 Hz. The controller tracks the VLA-generated targets while maintaining balance and completing the remaining whole-body motion.

\section{EVALUATION}

We conduct simulation and real-world experiments to evaluate whether HALOMI can effectively learn humanoid loco-manipulation with active perception from human demonstrations. Our evaluation is designed to address the following key questions:

\noindent\textbf{Q1: Loco-Manipulation Capability.}
Can HALOMI accomplish diverse loco-manipulation tasks through sparse head-hand interface?

\noindent\textbf{Q2: Human-to-Humanoid Alignment.}
Do ego-view alignment and controller-aware reference trajectory adaptation improve human-to-humanoid transfer?

\noindent\textbf{Q3: Active Perception.}
What role does active perception play in learning humanoid loco-manipulation from human demonstrations?

\noindent\textbf{Q4: Generalization.}
Can HALOMI generalize beyond the demonstrated settings to unseen object positions/appearances?

\subsection{Experimental Setup}
We evaluate HALOMI in two parts: controller execution and the entire system for loco-manipulation. We first perform simulation and real-world experiments to test the whole-body controller, evaluating whether it can achieve precise and robust head-hand tracking. 

We then evaluate HALOMI on five representative real-world humanoid loco-manipulation tasks, as shown in Fig.~\ref{fig:real_world_tasks}. For \textit{Bag Transfer}, \textit{Pick Bread and Place}, and \textit{Transfer Towel to Basket}, we train policies using ${\text{102}}$, ${\text{95}}$, and ${\text{96}}$ human demonstrations, respectively, and conduct quantitative real-world evaluations with 20 rollouts per setting. 
Extensive ablations and generalization tests are further conducted to analyze how key components support learning humanoid loco-manipulation from human demonstrations. 
For \textit{Squat-and-Grasp} and \textit{Tossing}, we provide qualitative analysis to evaluate capabilities beyond standard pick-and-place, including whole-body posture adaptation and dynamic motions.

\subsection{Can HALOMI Achieve Diverse Loco-Manipulation Tasks Through Sparse Head-Hand Interface?} 

\begin{figure}[htbp]
    \centering
    \includegraphics[width=\columnwidth]{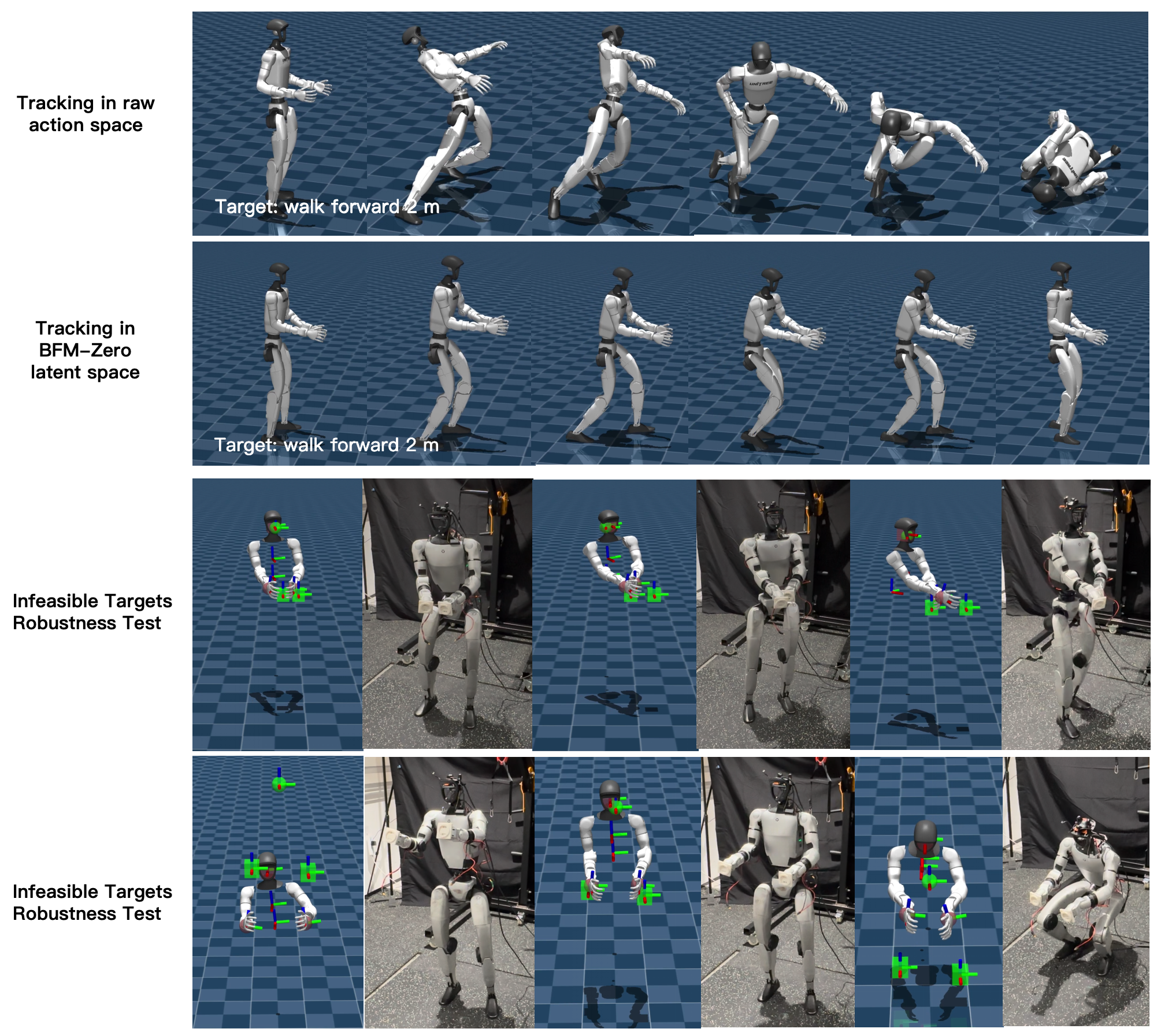}
    \vspace{-4.5mm}


    \vspace{-2mm}
     \caption{\textbf{Controller Robustness Evaluation.}
    Top row: raw action-space tracking loses balance under OOD target commands.
    Below: the manifold-constrained controller maintains stable and feasible whole-body behaviors under challenging OOD commands.
}
    \label{fig:loco_robustness}
    \vspace{-2mm}
\end{figure}

The controller executes the sparse head-hand targets predicted by the high-level VLA and completes the remaining whole-body motion. Therefore, we first test its tracking performance and robustness.
As shown in Table~\ref{tab:target_refinement}, we replay collected human head-hand trajectories that were not used for controller training and report the tracking errors. We use \(E_{\mathrm{MGBP}}\) to measure the average global tracking error over the head-hand target points. We evaluate three tasks that require accurate head-hand tracking for successful execution and cover diverse motion patterns.
We further evaluate the controller under sudden large target changes and infeasible motions. As shown in Fig.~\ref{fig:loco_robustness}, the controller maintains stable, non-aggressive behavior under these challenging commands.

\begin{table}[htbp]
\vspace{1mm}
\centering
\caption{\textsc{Tracking Performance.}}
\label{tab:target_refinement}
\scriptsize
\setlength{\tabcolsep}{2.5pt}
\begin{tabular}{lccc}
\toprule
Task
& \(E_{\mathrm{MGBP}}\) (m)
& \(E_{\mathrm{MGBP}}\) w/ Adapt. (m)
& Improvement (\%) \\
\midrule
Pick Bread and Place & 0.0436  & 0.0409 &  6.193 \\
Transfer Towel to Basket  & 0.0568  & 0.0523 &  7.923 \\
Squat-and-Grasp  & 0.1238  & 0.1163 &  6.058 \\
\bottomrule
\end{tabular}
 \vspace{-3mm}
\end{table}

From the perspective of the entire system, HALOMI is evaluated on five representative real-world loco-manipulation tasks. 
Specifically, Fig.~\ref{fig:real_world_tasks}(a) shows \textit{Bag Transfer}, which evaluates long-range navigation with active visual search. 
Fig.~\ref{fig:real_world_tasks}(b) shows \textit{Pick Bread and Place}, which evaluates moderately precise tabletop pick-and-place with active perception. Fig.~\ref{fig:real_world_tasks}(c) shows \textit{Transfer Towel to Basket}, which evaluates bimanual manipulation, whole-body coordination, and active perception. 
Fig.~\ref{fig:real_world_tasks}(d) and Fig.~\ref{fig:real_world_tasks}(e) further show \textit{Tossing} and \textit{Squat-and-Grasp}, which evaluate dynamic motion and deep-squat pick-and-place behaviors, respectively.

For the three quantitatively evaluated tasks, HALOMI achieves success rates of \(90\%\), \(85\%\), and \(80\%\), on Bag Transfer, Pick Bread and Place, and Transfer Towel to Basket, respectively. In \textit{Tossing}, the robot performs a fast object-release motion toward the target region, while in \textit{Squat-and-Grasp}, the robot performs a deep squat to grasp a low-positioned object, stands up, and turns to place it in the bag. Together, these results show that HALOMI can achieve diverse loco-manipulation behaviors from sparse head-and-hand targets with natural and stable whole-body hand-eye coordination.

\subsection{Does Our Human Data Processing Improve Human-to-Humanoid Transfer?}

We investigate whether our human data processing improves human-to-humanoid transfer by reducing observation-space and action-execution gaps.

\begin{figure}[htbp]
    \centering
    \vspace{-3.5mm}
    \includegraphics[width=\linewidth]{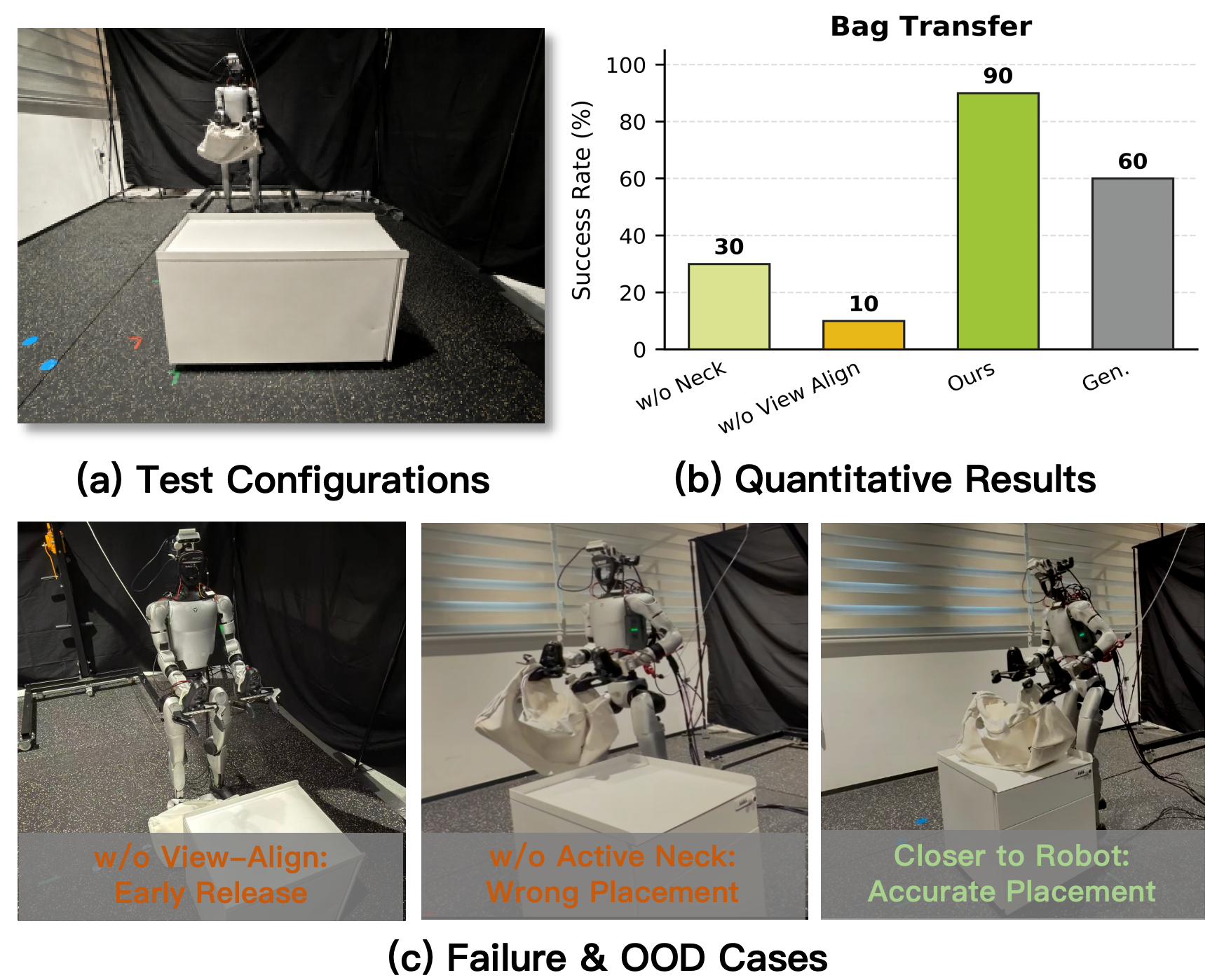}
    \caption{
    \textbf{Bag Transfer Task.}
    (a) Test scenarios with varied cabinet placements.
    (b) Quantitative results under ablation and generalization settings.
    (c) Representative failure and OOD cases across different settings.
    }
    \label{fig:bag_results}
    \vspace{-2mm}
\end{figure}

\noindent\textbf{Ego-view alignment.}
The ablation on \textit{Bag Transfer} shows the importance of ego-view alignment. As shown in Fig.~\ref{fig:bag_results}(b), removing ego-view alignment decreases the success rate from \(90\%\) to \(10\%\).
Without ego-view alignment, we observe that the robot can still approach the target but often stops at a constant offset from the desired placement region. This suggests that the human-to-humanoid ego-viewpoint gap biases the learned visual-action mapping, leading to systematic errors in ego-view guided manipulation.

\begin{figure}[htbp]
    \centering
    \includegraphics[width=\linewidth]{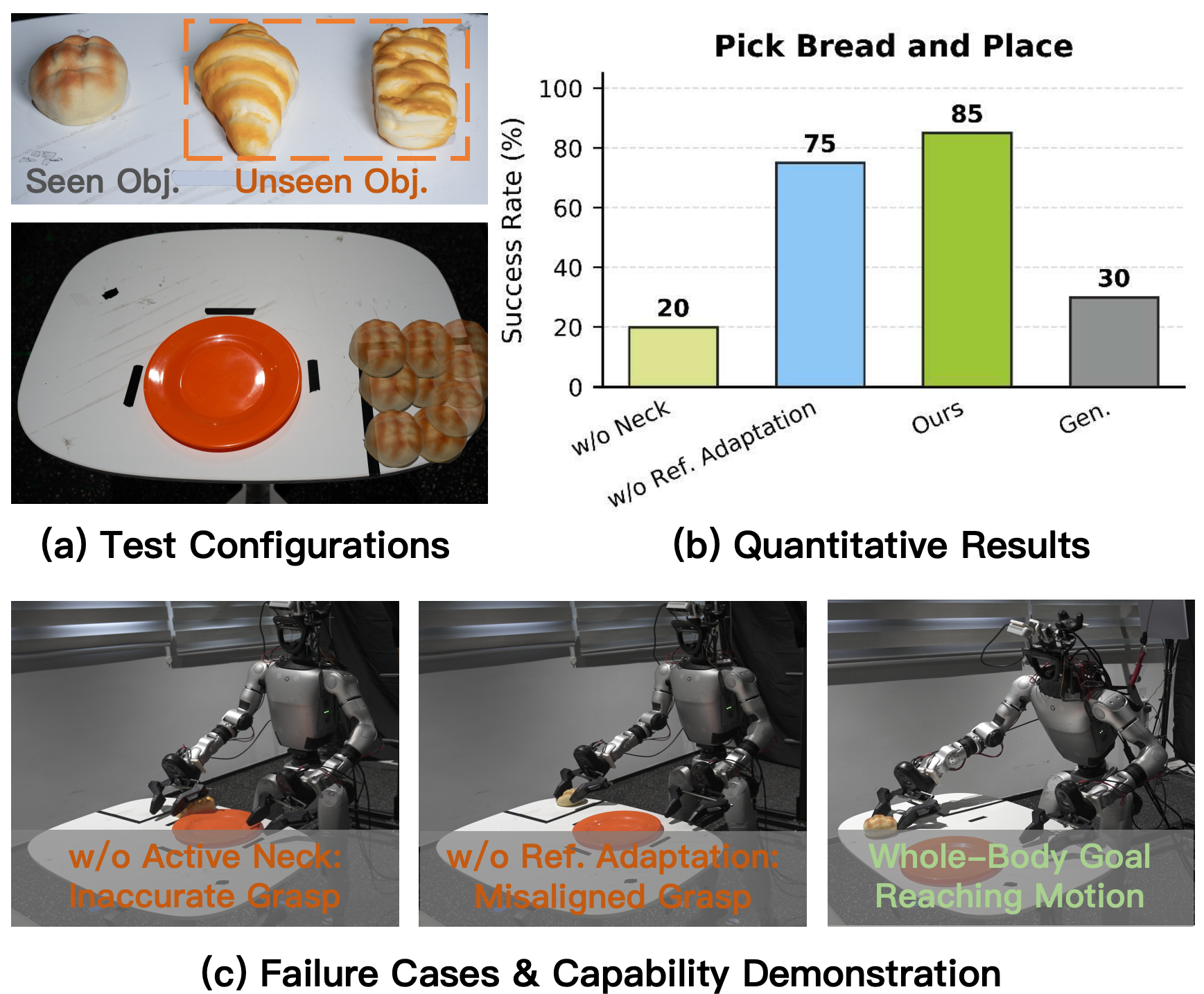}
    \caption{
    \textbf{Pick Bread and Place Task.}
    (a) Test scenarios with varied bread and plate placements.
    (b) Quantitative results under ablation and generalization settings.
    (c) Representative failure and capability cases across different settings.}
    \label{fig:bread_results}
     \vspace{-2mm}
\end{figure}

\noindent\textbf{Controller-aware reference trajectory adaptation.}
The effect of controller-aware reference trajectory adaptation is first quantified through tracking-error analysis.
As shown in Table~\ref{tab:target_refinement}, reference adaptation consistently reduces tracking errors across all three tasks, yielding an average error reduction of approximately \(6.725\%\).

Its downstream effect on human-to-humanoid transfer is further evaluated through task success rate and execution quality.

\noindent\textit{Pick Bread and Place.}
As shown in Fig.~\ref{fig:bread_results}(b), adding controller-aware reference trajectory adaptation improves the success rate from \(75\%\) to \(85\%\). Among successful trials, we observe improved execution accuracy: the robot grasps closer to the bread center and places the bread closer to the intended plate region.

\begin{figure}[h]
    \centering
    \includegraphics[width=\linewidth]{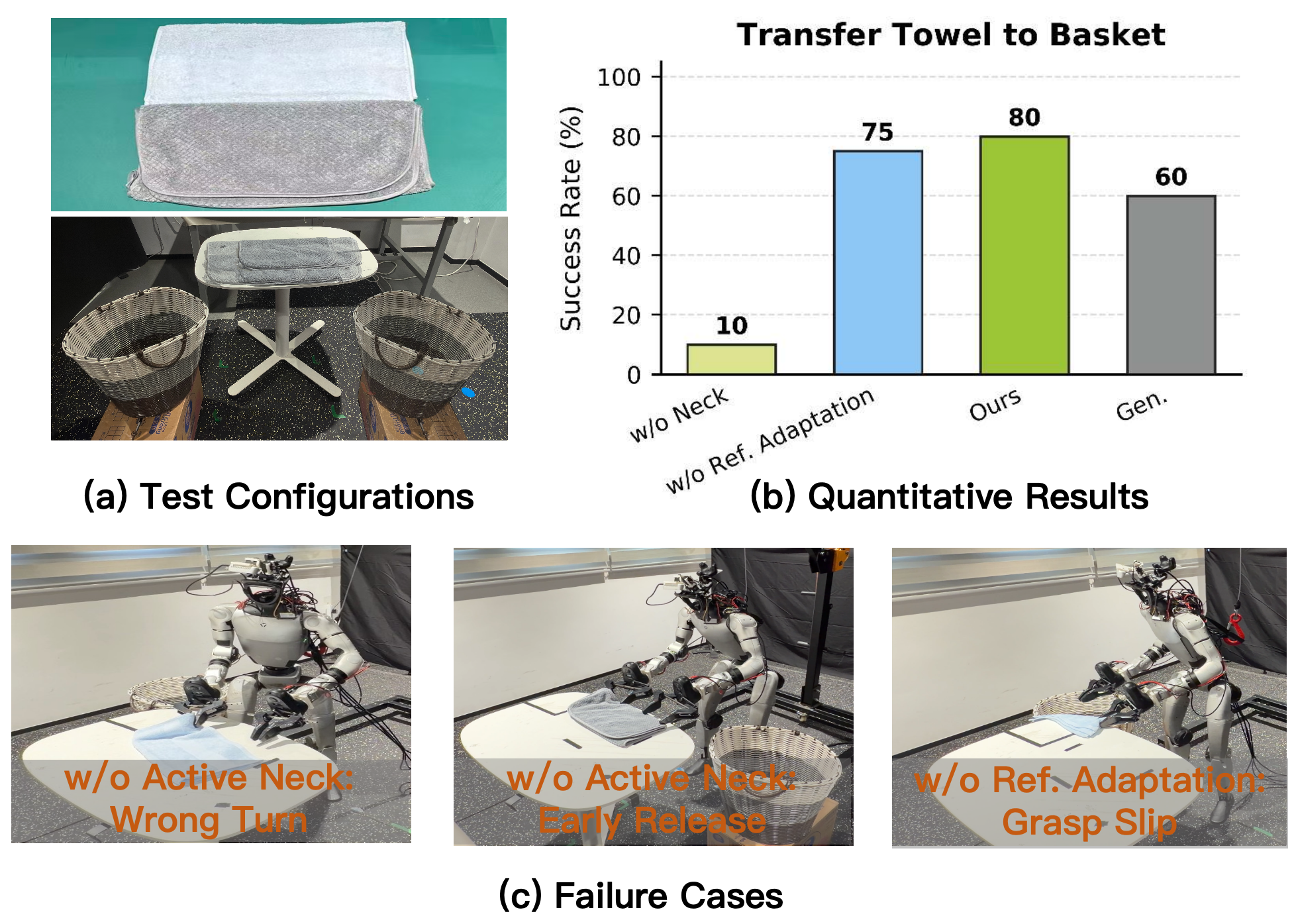}
    \caption{
    \textbf{Transfer Towel to Basket} (a) Test scenarios with varied towel and basket placements.
    (b) Quantitative results under ablation and generalization settings.
    (c) Representative failure cases across different settings.}
    \label{fig:towel_results}
     \vspace{-4.5mm}
\end{figure}

\noindent\textit{Transfer Towel to Basket.}
As shown in Fig.~\ref{fig:towel_results}(b), adaptation improves the success rate from \(75\%\) to \(80\%\). Beyond success rate improvement, qualitative results show more accurate grasps and smoother whole-body execution. The smoother execution also leads to steadier ego-view observations during transport and placement. 

\subsection{What Is the Role of Active Perception in Human-to-Humanoid Transfer?}
To evaluate the role of active perception in human-to-humanoid transfer, we ablate active neck control in two settings: tasks that require active viewpoint changes and tasks that can be completed from a static view.


\noindent\textbf{Tasks requiring active perception.}
In \textit{Bag Transfer} and \textit{Transfer Towel to Basket}, active perception is required to maintain task-relevant visual context across execution stages. 

For \textit{Bag Transfer}, as shown in Fig.~\ref{fig:bag_results}(b), disabling active neck control reduces the success rate to \(30\%\). Most failures occur when the cabinet is not in the center: as the robot walks forward, the cabinet gradually moves out of the central field of the ego-view, causing the policy to lose target localization.

For \textit{Transfer Towel to Basket}, as shown in Fig.~\ref{fig:towel_results}(b), disabling active neck control reduces the success rate from \(80\%\) to \(10\%\). We observe that the robot can often still approach and grasp the towel, but fails in the subsequent placement stage because it cannot actively search for the basket and bring it into the center of the head-camera view. 
These indicate that active head-view control is particularly important for tasks where the task-relevant visual focus changes across execution stages.

\noindent\textbf{Tasks solvable with a static view.}
In \textit{Pick Bread and Place}, a static ego-view can provide sufficient visual coverage for task completion. However, during human data collection, demonstrators naturally coordinate head and hand motions by shifting their gaze between the grasping and placement phases.
As shown in Fig.~\ref{fig:bread_results}(b), disabling active head control still reduces the success rate from \(85\%\) to \(20\%\). This suggests that active perception is not only necessary to enlarge the visual field, but also crucial to preserve the hand-eye coupling in human demonstrations.

\subsection{Generalization Ability of HALOMI}
We further evaluate whether HALOMI generalizes beyond the demonstrated settings under the representative distribution shifts: unseen spatial configurations and unseen object appearances.

\begin{figure}[h]
    \centering
    \vspace{-3.5mm}
    \includegraphics[width=\linewidth]{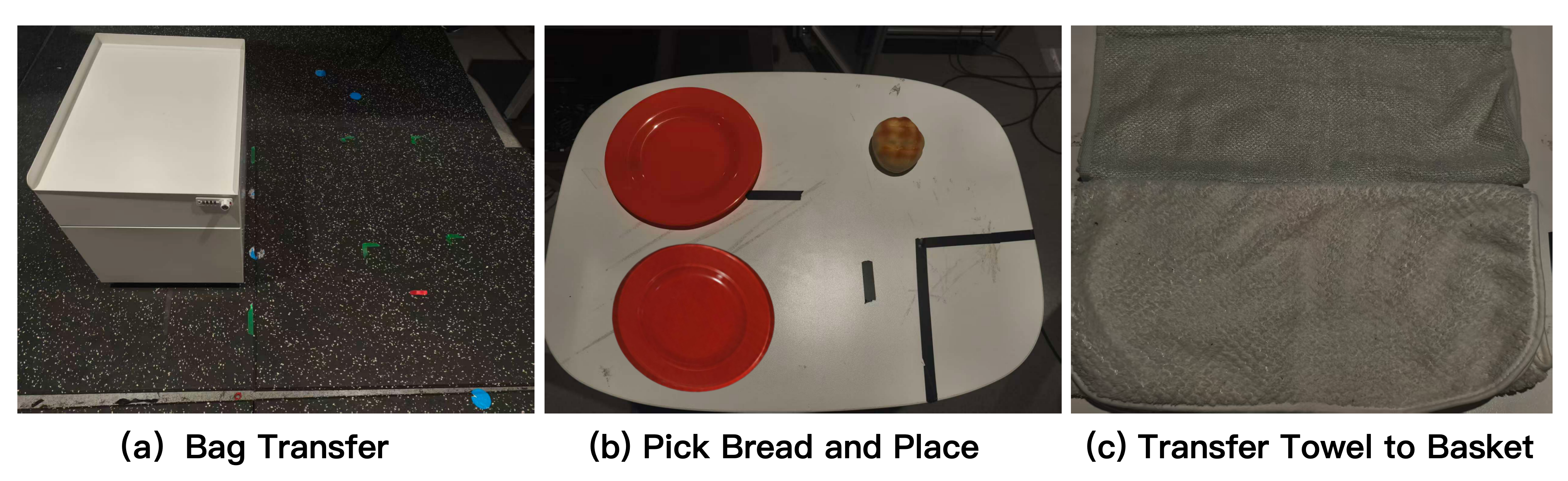}
    \caption{
    \textbf{OOD Test Configurations} 
    (a) Bag Transfer.
    (b) Pick Bread and Place.
    (c) Transfer Towel to Basket.}
    \label{fig:gen_settting}
     \vspace{-2.5mm}
\end{figure}

\noindent\textbf{Unseen Object Appearances.}
For \textit{Transfer Towel to Basket}, as shown in Fig.~\ref{fig:gen_settting}(c), we evaluate novel towels with appearances different from those in the demonstrations. The policy achieves a \(60\%\) success rate in the generalization setting, suggesting that HALOMI retains object-level visual generalization.

\noindent\textbf{Unseen Spatial Configurations.}
For \textit{Bag Transfer}, we evaluate the cabinet at unseen locations, as shown in Fig.~\ref{fig:gen_settting}(a). The policy maintains a \(60\%\) success rate under OOD cabinet placements, suggesting that our relative head-hand pose interface enables generalization to unseen target locations beyond the demonstration settings.

For \textit{Pick Bread and Place}, we further stress-tested this by introducing two types of spatial shifts, as illustrated in Fig.~\ref{fig:gen_settting}(b): changing the absolute bread/plate positions while preserving their relative layout, and changing both the absolute positions and the relative bread-to-plate layout.
The policy remains robust to global translations, achieving 6/10 success when the relative bread-to-plate arrangement is preserved. 
As shown in Fig.~\ref{fig:bread_results}(c), the robot can synthesize whole-body coordination that is not explicitly demonstrated, enabling the robot to reach far-away targets beyond the demonstrated workspace.
However, under bread-to-plate relative layout changes, the success rate dropped to 0/10. In these failures, the robot often grasps the bread but fails during placement, since the hand-target trajectory required to complete the placement is not covered by the demonstrated head-hand dataset.

\section{CONCLUSIONS}

We present HALOMI, a system for learning whole-body humanoid loco-manipulation with active perception directly from human demonstrations. We couple the robot-free egocentric demonstration interface with an automated human data processing pipeline, providing a scalable route to collect human data and bridge the human-to-humanoid embodiment gap. For robot execution, we integrate a humanoid platform with an active neck and a manifold-constrained whole-body controller, enabling robust world-frame tracking of sparse head-hand targets. Real-world experiments demonstrate that HALOMI enables versatile and challenging long-horizon humanoid loco-manipulation tasks with active perception. 


\addtolength{\textheight}{-12cm}   







\bibliographystyle{IEEEtran}
\bibliography{references}

\end{document}